\documentclass[10pt, conference, compsocconf]{IEEEtran}

\usepackage{graphicx}
\usepackage{amsmath,bm}
\usepackage{url}
\usepackage{multirow}
\usepackage{array}
\usepackage{tabularx,booktabs}
\usepackage{float}
\newcolumntype{C}{>{\centering\arraybackslash}X}
\DeclareMathOperator{\diag}{diag}
\DeclareMathOperator{\Poisson}{Poisson}

\begin{document}

\title{Convolutional Neural Networks Regularized by Correlated Noise}

\author{\IEEEauthorblockN{Shamak Dutta, Bryan Tripp}
\IEEEauthorblockA{Systems Design Engineering \& Center for Theoretical Neuroscience\\
University of Waterloo\\
Waterloo, Canada\\
\{s7dutta,bptripp\}@uwaterloo.ca}
\and
\IEEEauthorblockN{Graham W.~Taylor}
\IEEEauthorblockA{School of Engineering, University of Guelph\\
Vector Institute for Artificial Intelligence\\
Canadian Institute for Advanced Research\\
gwtaylor@uoguelph.ca}
}

\maketitle

\begin{abstract}
  Neurons in the visual cortex are correlated in their variability. The presence
  of correlation impacts cortical processing because noise cannot be averaged
  out over many neurons. In an effort to understand the functional purpose of
  correlated variability, we implement and evaluate correlated noise models in
  deep convolutional neural networks. Inspired by the cortex, correlation is
  defined as a function of the distance between neurons and their selectivity.
  We show how to sample from high-dimensional correlated distributions while
  keeping the procedure differentiable, so that back-propagation can proceed as
  usual. The impact of correlated variability is evaluated on the classification
  of occluded and non-occluded images with and without the presence of other
  regularization techniques, such as dropout. More work is needed to understand
  the effects of correlations in various conditions, however in 10/12 of the
  cases we studied, the best performance on occluded images was obtained from a
  model with correlated noise.
\end{abstract}

\begin{IEEEkeywords}
  Correlated Variability, Convolutional Neural Networks, Regularization, Stochastic Neurons
 \end{IEEEkeywords}

\IEEEpeerreviewmaketitle

\section{Introduction}

Convolutional neural networks (CNN) trained for object recognition tasks are
similar to the visual cortex in many ways. For example, early layers show Gabor-like receptive fields similar to V1 \cite{Zeiler2013}, late layers that are highly
predictive of V4, and inferior temporal cortex (IT) responses \cite{Yamins2014}.
However, these networks lack the correlated variability of neuron responses in
the human brain, among other major differences. In this paper, we discuss
methods to incorporate correlated variability into deep convolutional networks
and analyze its effect on recognition performance.

Studying stochastic neurons is interesting because the effect of stochasticity
on learning and computation in artificial neural systems may help us in modeling
biological neurons. In population coding schemes in the brain, the joint
activities of many neurons encode the value of a quantity. One advantage of
population coding is that the noise can be averaged out over many neurons.
However, the observation of correlation in spike variability
\cite{Zohary1994} is concerning because it can prevent noise averaging and
strongly affect cortical processing. The function of correlated variability in the brain is unclear. Our goal is to understand if correlated variability has benefits that can be realized in artificial neural networks by studying its
influence on certain tasks. One example of such a task is the long-standing
problem of object recognition under partial occlusion. This requires the ability
to robustly model invariance to certain transformations of the input. The
primary method of gaining invariance to transformations is data driven, either
by attempting to collect instances of the object under a wide variety of
conditions, or augmenting the existing dataset via synthetic transformations.
The authors in \cite{Wang2017} suggest that the right way to learn invariance
is not by adding
data, as occlusions follow a long-tail distribution which cannot be covered,
even by large-scale efforts in collecting data. This motivates the need to
modify the structure of the network to learn invariance. We hypothesize that
correlated variability might improve recognition performance in the challenging
setting of partial occlusion.

A common approach for regularizing deep neural networks is to inject noise during
training; for example, adding or multiplying noise to the hidden units of the neural network, like in dropout~\cite{Srivastava2014}. Most solutions include additive or multiplicative independent noise
in
the hidden units. This is widely used because of its simplicity and
effectiveness. We are motivated to consider correlated noise that is dependent
on the weights of the network, such as our proposal to add noise sampled from a
correlated distribution, where the correlation is a function of the differences in spatial position and selectivity of neurons, similar to the visual cortex.
One of the major concerns of stochasticity in neural networks is the tendency
to break differentiability, which prevents gradient computation via
back-propagation. Depending on the noise distribution, one may simply ignore
stochasticity in back-propagation (e.g.~the straight-through estimator in the case of
binary stochastic neurons~\cite{Bengio2013}), or one may apply computationally convenient
estimators for non-differentiable modules.
For the latter case, there has been
much work focused on back-propagation through non-differentiable
functions. For example, the re-parameterization trick in \cite{Kingma2013} allows
for
back-propagation through sampling from a broad class of continuous
distributions. Re-parameterization using the Gumbel-max trick and the softmax
function in
\cite{Maddison2016,Jang2016} allows for back-propagation through samples from
categorical discrete distributions. Advancements in the area of back-propagation
through non-differentiable modules is relevant to our work because it allows us
to consider many interesting types of noise models.

In this paper, we discuss four different types of noise models: independent
Gaussian, correlated Gaussian, independent Poisson, and correlated Poisson
(Section \ref{noiseFramework}). In the case of correlation, we also describe
how to construct correlation between samples (Section
\ref{correlationStructure}). Finally, we evaluate these different noise models
on the classification of occluded and non-occluded images.

\section{Related Work}

\textbf{Independent Noise Models:} The analysis of noise in deep networks has
focused on
models that use
independent noise to perturb the activations of neurons. For example, dropout \cite{Srivastava2014} is an effective method to
regularize neural networks where each unit is independently dropped with a
certain probability. Independent Gaussian noise has also been extensively
explored in \cite{Poole2014}, where noise is added to the input, or before or after the
non-linearity in other layers. The authors connect different
forms of independent noise injection to certain forms of optimization penalties
in a special form of a denoising autoencoder. An observation to note is that
additive Gaussian noise with zero-mean and variance equal to the unit activation relates to a penalty that encourages sparsity on the hidden unit
activations. This is interesting because the time intervals between spikes in a
cortical neuron are irregular and can be modeled using a Poisson process, such that the variance of the spike count in a small time interval is equal to the mean spike count. This motivates us to investigate whether we can model
artificial neurons as samples from a Poisson distribution. The work done in
\cite{Poole2014} focused on unsupervised learning in autoencoders and used the
learned hidden representations as inputs to a classifier. Our approach differs
in the fact that we are directly using a supervised classifier (CNN) to analyze the injection of noise.

\textbf{Back-propagation Through Random Nodes:} Gradient-based
learning which leverages back-propagation in neural networks requires that all operations that
depend on the trainable parameters be differentiable. Stochastic neural networks
usually involve samples from a distribution on the forward pass. However, the
difficulty is that we cannot back-propagate through the sampling operation. It is shown in
\cite{Bengio2013} that in the case of injection of additive or multiplicative
noise in a
computational graph, gradients can be computed as usual. We use this method in
Section~\ref{independentGaussianNoise}. Bengio et al.~\cite{Bengio2013} also introduce the concept of
a straight-through estimator, where a copy of the gradient with respect to the
stochastic output is used directly as an estimator of the gradient with respect
to the sigmoid (or any non-linearity) operator. Similar to the
straight-through estimator, in Sections
\ref{independentPoissonNoise} and \ref{correlatedPoissonNoise}, we use the expected
value of the Poisson distribution as if it were the output of the neuron during
back-propagation. The work in \cite{Maddison2016,Jang2016} allows for
back-propagation through samples from discrete categorical distributions. While
we do not use this technique in our work, it is possible to choose an upper
bound \(K\) to convert a Poisson distribution to a categorical distribution of
size \(K\). The sample from the distribution would be the expected value of
this estimated categorical distribution and back-propagation can proceed since
the entire process is made differentiable.

\textbf{Augmentation in Feature Space:} Dataset augmentation is a cheap and
effective way to generate more training data with variability that is expected
at inference time. Recently, some works have considered augmentation
techniques, such as the addition of noise, as well as interpolation and extrapolation from pairs of examples, not in input space, but in a learned feature space \cite{DeVries2017,Zhang2017-yx}. However,
\cite{DeVries2017} shows that simple noise models (e.g.~Gaussian) do not work
effectively when compared to extrapolation. We are motivated by the fact
that more sophisticated noise models (e.g.~correlated) could be useful for
feature space-based augmentation.

\section{Methods}

\subsection{Noise Framework} \label{noiseFramework}

We analyzed different types of independent and correlated noise to elucidate their
effects on neural networks. This helped us understand the impact of injecting
correlated noise when compared to the common practice of injecting independent noise.

\subsubsection{Independent Gaussian Noise} \label{independentGaussianNoise}

We consider \(\mathrm{h_{i}}\) to be the output of a stochastic neuron. The output is a
deterministic function of a differentiable transformation \(a_{i}\) and a noise
source \(\mathrm{z}_{i}\), as considered in \cite{Bengio2013}. \(a_{i}\) is typically a
transformation of its inputs (vector output of other neurons) and trainable
parameters (weights and biases of a network). The output of the stochastic
neuron is:

\begin{equation} \label{eq:1}
	\mathrm{h_{i}} = f(a_{i}, \mathrm{z}_{i}).
\end{equation}

As long as \(\mathrm{h_{i}}\) is differentiable with respect to \(a_{i}\) and has a
non-zero gradient, we can train the network with back-propagation. In this section,
one form of noise we consider is additive independent Gaussian noise with zero
mean and \(\sigma^2\) variance, where:

\begin{equation} \label{eq:2}
	f(a_{i}, \mathrm{z}_{i}) = a_{i} + \mathrm{z}_{i},
\end{equation}
\begin{equation} \label{eq:3}
	\mathrm{z}_{i} \sim \mathcal{N}(0, \sigma^{2}).
\end{equation}

At test time, we can compute the expectation of the noisy activations by
sampling from the distribution \(\mathcal{N}(a_{i}, \sigma^{2})\); however, this can be
computationally expensive for large datasets. Instead, we can approximate the
expectation by scaling the units by their expectation, as in dropout~\cite{Srivastava2014}. Since we are using zero mean additive noise, no scaling
is required, as the expectation does not change.

A special case of Equation~\ref{eq:3} is when \(\sigma^{2} = a_{i}\). The distribution
of activations for a specific stimulus will follow \(\mathcal{N}(a_{i}, a_{i})\),
which has a Fano factor of 1. This is similar to biological neurons, which
exhibit Poisson-like statistics with a Fano factor of approximately 1. It also
means that \(\mathrm{z}_{i}\) is now a function of \(a_{i}\) and the gradient of \(\mathrm{z}_{i}\)
with respect to \(a_{i}\) exists through re-parameterization. A sample from a
normal distribution \(\mathcal{N}(\mu, \sigma^{2})\) can be constructed as:

\begin{equation} \label{eq:4}
	\begin{split}
		\mathrm{x}_{i} \sim \mathcal{N}(0, 1), \\
		\mathrm{z}_{i} = \mu + \sigma \mathrm{x}_{i}.
	\end{split}
\end{equation}

In the case when \(\sigma^{2} = a_{i}\) and
\(\mu = 0\), \(\mathrm{z}_{i} = \sqrt{a_{i}} \mathrm{x}_{i}\). In practice, the gradients can be unstable for the square-root
function for small values. To solve this problem, we add a small value of
\(\epsilon = 0.0001\) to the argument of the square-root function.

\subsubsection{Correlated Gaussian Noise} \label{correlatedGaussianNoise}

We consider \(\mathbf{h} \in {\rm I\!R^{n}}\) to be a vector of outputs of \(n\)
stochastic neurons. \(\mathbf{h}\) is a sum of a noise vector \(\bm{z} \in {\rm
I\!R^{n}}\) and a vector output \(\bm{a} \in {\rm I\!R^{n}}\), which is a
differentiable transformation of its inputs and trainable parameters, similar to
Section~\ref{independentGaussianNoise}. The vector output is:

\begin{equation} \label{eq:5}
 \mathbf{h} = \mathbf{z} + \bm{a},
\end{equation}
\begin{equation} \label{eq:6}
  \mathbf{z} \sim \mathcal{N}(\bm{0}, \bm{\Sigma}).
\end{equation}

Given a desired mean \(\bm{\mu}\) and correlation matrix \(\bm{\Sigma}\), \(
\mathbf{z} \in {\rm I\!R^{n}}\) is sampled as follows:

\begin{enumerate}
  \item[--] Sample \(\mathbf{X} \sim \mathcal{N}(\bm{0}, \bm{I}_{n})\)
  \item[--] Compute the Cholesky decomposition: \(\bm{\Sigma}\) = \(LL^{*}\), where \(L^{*}\) is the conjugate transpose of \(L\).
  \item[--] \(\mathbf{z} = \bm{\mu} + L\mathbf{X} \)
  \item[--] If \(\bm{\sigma} \in {\rm I\!R^{n}}\) is the desired standard
  deviation, then
    \begin{equation} \label{eq:7}
      \mathbf{z} = \diag(\bm{\sigma}) (\bm{\mu} + L\mathbf{X}),
    \end{equation}
    where \(\diag(\bm{\sigma}) \in {\rm I\!R^{n \times n}}\) is a matrix with
    the standard deviations on the diagonal and zeros elsewhere.
\end{enumerate}

As described in Section~\ref{tuningSimilarity}, the correlation matrix
\(\bm{\Sigma}\) is a function of the trainable weights of the network. This
implies that
the entire sampling process is differentiable with respect to the parameters
\(\mu\), \(\sigma\), and the trainable weights, and that gradient-based learning
can proceed as
usual.

\subsubsection{Independent Poisson Noise} \label{independentPoissonNoise}

We consider \(\mathrm{h_{i}}\) to be the output of a stochastic neuron. The definition of \(a_{i}\) is the same as in
Section~\ref{independentGaussianNoise}. In the case of independent Poisson
noise, the output of the neuron is:

\begin{equation} \label{eq:8}
  \mathrm{h}_{i} \sim \Poisson(\lambda = a_{i}),
\end{equation}

\noindent where the mean is given by \(a_{i}\). This poses a problem in back-propagation, as there is no gradient of \(\mathrm{h}_{i}\)
with respect to the parameter \(\lambda = a_{i}\). The re-parameterization trick
in \cite{Kingma2013} cannot be applied in this case because the Poisson distribution is
discrete. To avoid this problem, we set the output of the unit to its mean value during the backward pass,
which is \(a_{i}\). We still propagate the sample from the distribution on the
forward pass. This is similar to the straight-through estimator for
back-propagation through stochastic binary neurons \cite{Bengio2013}. The
Gumbel-softmax method \cite{Maddison2016,Jang2016} can be used here if the Poisson
distribution is
converted to
a categorical distribution using an upper threshold \(K\). We plan to explore
this avenue in future work.

\subsubsection{Correlated Poisson Noise} \label{correlatedPoissonNoise}

Similar to Section~\ref{correlatedGaussianNoise}, \(\mathbf{h} \in {\rm I\!R^{n}}\)
is a vector of outputs of \(n\) stochastic neurons. \(\mathbf{h}\) is a sample from
a correlated Poisson distribution of mean \(\bm{\lambda}\) and correlation
matrix \(\bm{\bm{\Sigma}}\),

\begin{equation} \label{eq:9}
  \mathbf{h} \sim \Poisson(\bm{\lambda}, \bm{\Sigma}).
\end{equation}

We draw approximate samples from this distribution in the following way. Given a
desired mean \(\bm{\lambda} \in {\rm I\!R^{n}}\) and correlation matrix
\(\bm{\Sigma} \in {\rm I\!R^{n \times n}}\):

\begin{enumerate}
  \item[--] Sample \(\mathbf{X} \sim \mathcal{N}(\bm{\mu}, \bm{\Sigma})\), where \(\bm{\mu}\) is chosen arbitrarily
  \item[--] Apply the univariate normal cumulative distribution function (CDF): \(\bm{Y} = F_{x}(\mathbf{X}; \mu=0, \sigma=1)\)
  \item[--] Apply the quantile (inverse CDF) of the Poisson distribution: \(
  \mathbf{z} \sim F_{z}^{-1}(\bm{Y}; \bm{\lambda})\)
\end{enumerate}

This sampling procedure is non-differentiable for a correlated Poisson. As in
Section~\ref{independentPoissonNoise}, we estimate the gradient using the
mean \(\bm{\lambda}\) on the backward pass. The actual correlations between
Poisson variables depend on the rates \(\bm{\lambda}\) and are smaller than
the desired correlation matrix \(\bm{\Sigma}\).

\begin{figure}[ht]
\begin{center}
\includegraphics[width=\linewidth]{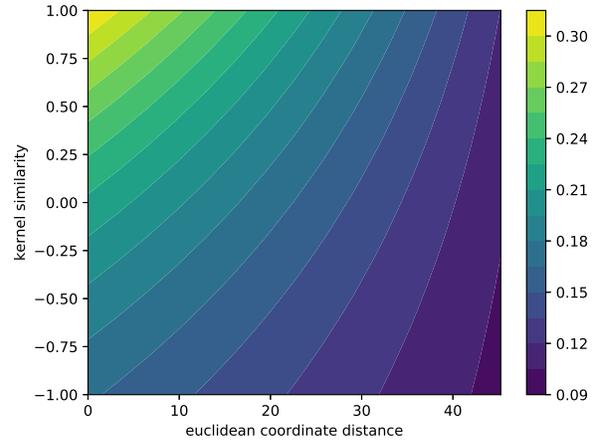}
\end{center}
\vspace{-5mm}
\caption{Correlation dependence on distance and tuning similarity for \(a = 0.225\), \(b=0.0043\), \(c=0.09\) and \(\tau=1.87\). The correlation value of the color plot is indicated by the bar on the right.}
\label{fig:heatmap}
\end{figure}

\subsection{Correlation Matrix} \label{correlationStructure}

In this section, we discuss how the correlation matrix \(\bm{\Sigma}\) is
constructed, which is used in Sections~\ref{correlatedGaussianNoise} and
\ref{correlatedPoissonNoise}. Note that the neuron equivalent in a CNN is a unit in the output feature map of a given layer.

Neurons in the cortex are correlated in their stochasticity. This correlation is
a function of the spatial spread and tuning similarity \cite{Smith2008}.
In the visual cortex, correlations have been studied between neurons in the same
visual area.
Analogously, we consider correlations between units in the same layer of a CNN.
The details of
spatial similarity and tuning
similarity are described in Sections~\ref{spatialSimilarity} and
\ref{tuningSimilarity}. For a given layer in a convolutional network, if the
width and height of the feature maps, as well as the number of feature maps are defined as
\(w\), \(h\), and \(k\), respectively, then the dimension \(d\) of the correlated
distribution we draw samples from is \(d = whk\) and the correlation matrix is
\(\bm{\Sigma} \in {\rm I\!R^{whk \times whk}}\).

Similar to a relationship suggested in \cite{Smith2008}, the correlation between
two neurons, \(x_{1}\) and \(x_{2}\), is determined as:

\begin{equation} \label{eq:10}
  f(x_{1}, x_{2}) = [a - b(d(x_{1}, x_{2}))]^{+} \cdot e^{\frac{k(x{1}, x{2}) - 1}{\tau}} + c,
\end{equation}

\noindent where \([\cdot]^{+}\) is \(\max(\cdot, 0)\), \(d(\cdot, \cdot)\) is a function
that returns the scaled Euclidean distance between two neurons, \(k(\cdot,
\cdot)\) is a function that returns the tuning similarity between two neurons
(bounded in \([-1,1]\)), and \(a\), \(b\), \(c\), and \(\tau\) are hyper-
parameters. The correlation is summarized in Figure~\ref{fig:heatmap} for
specific values of \(a\), \(b\), \(c\), and \(\tau\). To summarize, within a
layer of a CNN, every neuron is correlated to every
other neuron in the same layer by a value determined by how far apart they are
and their tuning similarity.

\subsubsection{Spatial Similarity} \label{spatialSimilarity}

The spatial similarity between two units in any output feature map within a layer is
determined by the Euclidean distance between the coordinates of the neurons
within their respective feature maps. The spatial distance between two neurons
that are a part of the \(i^{th}\) and \(j^{th}\) feature maps, respectively, with
coordinates \(\bm{p} = (x_{i}, y_{i})\) and \(\bm{q} = (x_{j}, y_{j})\)
is:

\begin{equation}
	d(\bm{p}, \bm{q}) = \sqrt{(x_{i} - x_{j})^2 + (y_{i} - y_{j})^2},
\end{equation}

Since the dimensions of the feature map do not change as training progresses, we
can pre-compute the spatial
correlations for all pairs of neurons before training begins.

\begin{table}[b!]
	\caption{Modified version of AllConvnet (ALL-CNN-C) architecture with 10 filters in layer 1 used with CIFAR-10.}
	\label{table:allconvnetArchitecture}
	\begin{center}
	 	\begin{tabular}{c}
		 \hline
		 Input: 32 \(\times\) 32 RBG image \\
		 \hline
		 Layer 1: 3 \(\times\) 3 conv. 10 filters, ReLU \\
		 Layer 2: 3 \(\times\) 3 conv. 96 filters, ReLU \\
		 Layer 3: 3 \(\times\) 3 conv. 96 filters, stride \(= 2\), ReLU \\
		 Layer 4: 3 \(\times\) 3 conv. 192 filters, ReLU \\
		 Layer 5: 3 \(\times\) 3 conv. 192 filters, ReLU \\
		 Layer 6: 3 \(\times\) 3 conv. 192 filters, stride \(= 2\), ReLU \\
		 Layer 7: 3 \(\times\) 3 conv. 192 filters, ReLU \\
		 Layer 8: 1 \(\times\) 1 conv. 192 filters, ReLU \\
		 Layer 9: 1 \(\times\) 1 conv. 10 filters, ReLU \\
		 Layer 10: global averaging over 6 \(\times\) 6 spatial dimensions \\
		 10-way softmax \\
		 \hline
		\end{tabular}
	\end{center}
\end{table}

\begin{table}[]
	\caption{Stochastic models evaluated as part of layer 1 tests.}
	\label{table:stochasticModels}
	\begin{center}
	 	\begin{tabular}{cc}
		 \hline
		 \textbf{Model name}&\textbf{abbreviation}\\
		 \hline
		 AllConvNet baseline &Baseline\\
		 AllConvNet ind. Gaussian \(\sigma = 1.0\) &IG\_A\\
		 AllConvNet ind. Gaussian \(\sigma = a_{i}\) &IG\_B\\
		 AllConvNet correlated Gaussian &CG\\
		 AllConvNet ind. Poisson &IP\\
		 AllConvNet correlated Poisson &CP\\
		 \hline
		\end{tabular}
	\end{center}
\end{table}

\subsubsection{Tuning Similarity} \label{tuningSimilarity}

The tuning similarity between two units in any output feature maps within a layer is
determined by the cosine similarity of the normalized weight matrices that
produced them. Consider a weight tensor (kernel in a convolutional network) of
dimension \(d = k \times k \times m \times n\), where \(k\) is the kernel size,
\(m\) is the number of input channels from the previous layer, and \(n\) is the
number of kernels. The \(i^{th}\) kernel \(\bm{w_{i}}\) is of dimension \(d =
kkm
\times 1\). The tuning similarity between two neurons in the \(i^{th}\) and
\(j^{th}\) feature maps is determined as:

\begin{equation}
	k(x_{i}, x_{j})= \left(\frac{\bm{w_{i}}}{\lVert \bm{w_{i}} \rVert}\right)^{T} \cdot \frac{\bm{w_{j}}}{\lVert \bm{w_{j}} \rVert} ,\
\end{equation}

\noindent since we are calculating the cosine similarity, \(k(\cdot, \cdot) \in
[-1, 1]\). Note that the tuning similarity solely depends on the feature maps that
the output units are a part of.

\begin{table*}
 \caption{Recognition performance of layer 1 stochastic behavioural models with
 no additional regularization over 10 runs}
\label{table:recog1}
\begin{tabularx}{\textwidth}{@{}l*{15}{C}c@{}}
\toprule
Model & \multicolumn{2}{c}{Test Set} & \multicolumn{2}{c}{Central Occlusion} &
\multicolumn{2}{c}{Checker Board} & \multicolumn{2}{c}{Horizontal Bars} &
\multicolumn{2}{c}{Vertical Bars} & \multicolumn{2}{c}{Horizontal Half} &
\multicolumn{2}{c}{Vertical Half} \\ \midrule
 & mean (\%) & s.d. (\%) & mean (\%) & s.d. (\%) & mean (\%) & s.d. (\%) & mean
 (\%) & s.d. (\%) & mean (\%) & s.d. (\%) & mean (\%) & s.d. (\%) & mean (\%)
 & s.d. (\%) \\
Baseline & 75.5 & 0.2 & 62.1 & 0.1 & 46.8 & 0.3 & 35.7 & 0.3 & 34.8 &
0.4 & 41.1 & 0.4 & 40.3 & 0.3 \\
IG\_A & 72.6 & 0.3 & 61.6 & 0.2 & 45.5 & 0.2 & 36.1 & 0.3 & 33.0 &
0.2 & 43.0 & 0.3 & 39.5 & 0.3 \\
IG\_B & 75.9 & 0.1 & 63.3 & 0.1 & 49.6 & 0.2 & 41.3 & 0.4 & 35.1 &
0.4 & 40.6 & 0.6 & 37.2 & 0.3 \\
CG & \textbf{80.8} & 0.1 & \textbf{69.1} & 0.1 & \textbf{58.1} & 0.2 &
\textbf{51.5} & 0.4 & \textbf{40.6} & 0.1 & \textbf{47.4} & 0.3 & 41.6 &
0.3 \\
IP & 71.8 & 0.8 & 59.6 & 0.5 & 44.6 & 0.6 & 35.2 & 0.7 & 32.7 & 0.4 &
40.2 & 0.4 & 36.3 & 0.4 \\
CP & 76.8 & 0.5 & 64.3 & 0.2 & 47.5 & 0.3 & 44.7 & 0.3 & 37.6 & 0.2
& 44.6 & 0.4 & \textbf{45.5} & 0.2 \\ \bottomrule
\end{tabularx}
\end{table*}

\begin{table*}
 \caption{Recognition performance of layer 1 stochastic behavioural models with
 dropout over 10 runs.}
\label{table:recog2}
\begin{tabularx}{\textwidth}{@{}l*{15}{C}c@{}}
\toprule
Model & \multicolumn{2}{c}{Test Set} & \multicolumn{2}{c}{Central Occlusion} &
\multicolumn{2}{c}{Checker Board} & \multicolumn{2}{c}{Horizontal Bars} &
\multicolumn{2}{c}{Vertical Bars} & \multicolumn{2}{c}{Horizontal Half} &
\multicolumn{2}{c}{Vertical Half} \\
\midrule
 & mean (\%) & s.d. (\%) & mean (\%) & s.d. (\%) & mean (\%) & s.d. (\%) &
 mean (\%) & s.d. (\%) & mean (\%) & s.d. (\%) & mean (\%) & s.d. (\%) & mean (
 \%) & s.d. (\%) \\
Baseline & 84.2 & 0.1 & 73.8 & 0.1 & 63.1 & 0.1 & 46.9 & 0.2 & 55.1 & 0.2 &
47.5 & 0.2 & 43.3 & 0.3 \\
IG\_A & \textbf{85.1} & 0.1 & 74.7 & 0.1 & 67.3 & 0.1 & 50.2 & 0.3 &
55.4 & 0.2 & \textbf{50.6} & 0.2 & 46.1 & 0.3 \\
IG\_B & 84.5 & 0.1 & 74.6 & 0.1 & 66.9 & 0.1 & 50.1 & 0.3 & 56.4 & 0.2 & 49.7
& 0.2 & 45.8 & 0.2 \\
CG & 84.2 & 0.1 & \textbf{75.5} & 0.1 & 58.4 & 0.2 & \textbf{54.1} & 0.2 &
57.5 & 0.2 & 47.4 & 0.2 & \textbf{47.2} & 0.2 \\
IP & 84.4 & 0.2 & 75.1 & 0.1 & 63.8 & 0.1 & 52.3 & 0.2 & \textbf{58.2} & 0.1 &
49.9 & 0.2 & 43.2 & 0.2 \\
CP & 84.9 & 0.2 & 75.4 & 0.1 & \textbf{68.1} & 0.3 & 53.1 & 0.3 & 57.0 & 0.1 &
47.9 &
0.2 & 44.6 & 0.3
\\
\bottomrule
\end{tabularx}
\end{table*}

\begin{figure}[t]
\begin{center}
\includegraphics[width=\linewidth]{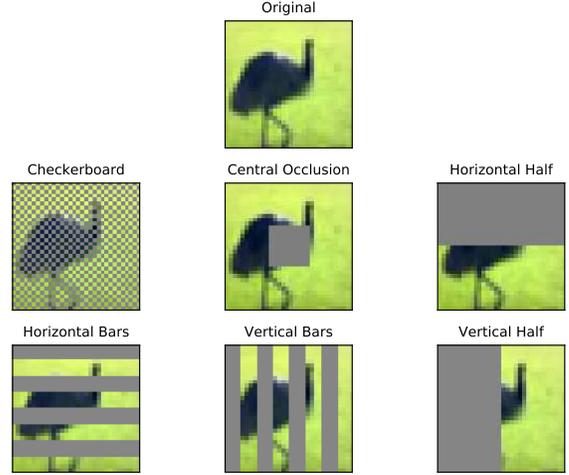}
\end{center}
\vspace{-0.5cm}
\caption{Different types of occlusions used to evaluate recognition performance. Images shown are taken from CIFAR-10.}
\label{fig:occlusions}
\end{figure}

\section{Experiments} \label{experiments}

We performed preliminary experiments with the aforementioned noise models using
an architecture equivalent to the AllConvnet network \cite{Springenberg2014}
(All-CNN-C) with one exception: the first layer contains 10 feature maps instead
of 96, as shown in Table~\ref{table:allconvnetArchitecture}. This was done for
computational tractability, as the correlation matrix grows as
\((whk)^{2}\), where \(w\) and \(h\) are the width and height of a feature map, and
\(k\) is the number of feature maps in a layer. As a result, the sampling
procedure from a correlated distribution can slow training by a large
amount. While we do not aim for state-of-the-art results, our goal is to obtain
a model that achieves respectable performance on CIFAR-10 and analyze the
effect of adding various kinds of noise. All models are trained for 100 epochs and the final
chosen model in each case is the one with the lowest validation loss during
training to prevent overfitting.

The CIFAR-10 dataset is used for all experiments, split into 50,000 training
images and 10,000 test images. The training set does not contain any occluded images. All experiments were performed using the
TensorFlow framework \cite{tensorflow2015}. We evaluate classification performance on the test
set, including occluded versions of the test set, as shown in Figure~\ref{fig:occlusions}.

We first experiment by incremental addition of the stochastic behaviour to
understand its effect. The baseline model is described in Table~\ref{table:allconvnetArchitecture}. First, we add noise in layer 1 of the
AllConvNet architecture. The different noise architectures for layer 1 are shown
in Table~\ref{table:stochasticModels}. The results of this experiment are shown in
Section~\ref{noDrop}.

We then analyze whether the benefits of noise can be realized in the
presence of another regularizer by
evaluating the effect of noise in the presence of dropout. In this case, the baseline model
shown in Table~\ref{table:allconvnetArchitecture} is trained with dropout in
specific layers. As before, we add different types of noise to layer 1 in order to
understand its impact, as described in Section~\ref{yesDrop}.

\begin{table*}
 \caption{Class Prediction Accuracy for (A) AllConvNet baseline model without dropout and (B) AllConvNet Correlated Gaussian model without dropout.}
\label{table:classPredict1}
\begin{tabularx}{\textwidth}{@{}l*{28}{C}c@{}}
\toprule
	Class     & \multicolumn{2}{c}{Test Set} & \multicolumn{2}{c}{Central Occlusion} & \multicolumn{2}{c}{Checker Board} & \multicolumn{2}{c}{Horizontal Bars} & \multicolumn{2}{c}{Vertical Bars} & \multicolumn{2}{c}{Horizontal Half} & \multicolumn{2}{c}{Vertical Half}\\

	& A & B & A & B & A & B & A & B& A & B & A & B & A & B \\
\midrule

airplane 	& 78.9 & 78.2 & 73.3 & 66.6 & 54.0 & 52.8 & 76.9 & 56.1 & 51.5 & 59.3 & 70.1 & 70.8 & 50.1 & 65.8 \\
automobile 	& 87.7 & 89.0 & 79.6 & 83.1 & 18.1 & 29.7 & 42.1 & 39.0 & 18.4 & 18.5 & 62.8 & 63.4 & 36.2 & 33.1 \\
bird 		& 69.4 & 73.2 & 57.7 & 58.7 & 46.2 & 51.9 & 32.1 & 38.2 & 52.8 & 57.1 & 30.5 & 35.8 & 79.5 & 76.0 \\
cat 		& 59.6 & 62.5 & 35.2 & 48.7 & 37.8 & 44.2 & 52.2 & 44.6 & 54.8 & 45.7 & 19.2 & 38.0 & 22.6 & 42.3 \\
deer 		& 79.6 & 82.2 & 68.8 & 63.4 & 63.9 & 69.7 & 63.1 & 73.6 & 50.4 & 50.7 & 59.7 & 58.8 & 44.1 & 31.8 \\
dog 		& 70.5 & 76.1 & 60.4 & 72.6 & 48.4 & 58.0 & 32.5 & 66.7 & 48.3 & 68.3 & 36.7 & 20.8 & 38.9 & 48.5 \\
frog 		& 85.9 & 88.2 & 74.5 & 75.7 & 64.0 & 71.4 & 15.0 & 42.1 & 16.9 & 23.1 & 60.2 & 67.0 & 45.2 & 32.3 \\
horse 		& 79.2 & 83.5 & 71.1 & 75.1 & 40.3 & 51.4 & 16.9 & 36.9 & 37.8 & 44.3 & 38.5 & 41.2 & 23.5 & 55.4 \\
ship 		& 88.6 & 87.8 & 71.9 & 73.1 & 70.0 & 70.4 & 51.2 & 73.6 & 53.8 & 47.1 & 84.8 & 81.8 & 73.8 & 53.1 \\
truck 		& 87.1 & 89.7 & 70.5 & 76.4 & 67.9 & 60.7 & 33.3 & 26.2 & 31.2 & 35.2 & 41.4 & 53.0 & 31.4 & 47.1 \\
\bottomrule
\end{tabularx}
\end{table*}

\begin{table*}
 \caption{Class Prediction Accuracy for (A) AllConvNet baseline model with
 dropout and (B) AllConvNet Correlated Poisson model with dropout.}
\label{table:classPredict2}
\begin{tabularx}{\textwidth}{@{}l*{28}{C}c@{}}
\toprule
	Class     & \multicolumn{2}{c}{Test Set} & \multicolumn{2}{c}{Central Occlusion} & \multicolumn{2}{c}{Checker Board} & \multicolumn{2}{c}{Horizontal Bars} & \multicolumn{2}{c}{Vertical Bars} & \multicolumn{2}{c}{Horizontal Half} & \multicolumn{2}{c}{Vertical Half}\\

	& A & B & A & B & A & B & A & B& A & B & A & B & A & B \\
\midrule

airplane    & 91.2 & 86.6 & 87.2 & 82.0 & 81.7 & 75.4 & 85.1 & 71.3 & 85.7 &
78.3 & 89.4 & 69.3 & 79.8 & 76.4 \\

automobile  & 92.5 & 93.1 & 80.6 & 82.9 & 38.0 & 49.5 & 9.5  & 20.3 & 13.8 &
36.7 & 56.3 & 62.8 & 19.8 & 34.2 \\

bird 		& 81.7 & 84.5 & 65.8 & 77.1 & 56.3 & 69.8 & 8.3  & 52.6 & 44.8 &
67.8 & 29.1 & 54.5 & 77.0 & 76.7 \\

cat         & 75.3 & 65.8 & 63.1 & 53.8 & 55.6 & 45.2 & 10.7 & 47.4 & 49.9 &
49.1 & 24.3 & 30.5 & 50.9 & 62.5 \\

deer        & 87.3 & 83.8 & 75.4 & 75.6 & 80.7 & 78.9 & 51.3 & 83.7 & 33.3 &
63.5 & 64.7 & 58.8 & 27.1 & 59.2 \\

dog         & 75.7 & 83.2 & 57.1 & 71.9 & 48.7 & 62.6 & 10.9 & 52.0 & 54.2 &
65.7 & 37.7 & 40.2 & 19.7 & 43.8 \\

frog        & 92.3 & 87.5 & 77.1 & 69.5 & 85.5 & 73.8 & 0.9  & 7.2 & 5.2  &
36.6 & 57.1 & 67.7 & 17.4 & 38.5 \\

horse       & 92.2 & 91.5 & 86.5 & 86.6 & 61.2 & 68.2 & 24.5 & 55.0 & 58.1 &
64.6 & 54.2 & 55.2 & 24.5 & 46.7 \\

ship        & 93.0 & 92.0 & 82.4 & 79.9 & 79.4 & 82.8 & 58.6 & 73.4 & 67.1 &
62.7 & 75.1 & 88.0 & 78.5 & 64.9 \\

truck 		& 90.9 & 90.3 & 86.3 & 85.5 & 63.9 & 74.4 & 18.6 & 43.5 & 30.0 &
58.3 & 58.9 & 63.3 & 10.3 & 50.4 \\

\bottomrule
\end{tabularx}
\end{table*}

\subsection{Absence of Dropout} \label{noDrop}

The classification performance of the models that do not incorporate dropout are
summarized in Table~\ref{table:recog1}. The models are abbreviated, as shown in
Table~\ref{table:stochasticModels}.

Independent Gaussian noise does not seem to have an appreciable effect on the
learning of the network. We varied \(\sigma\) in the set \(\{0.1, 0.5, 1.0,
1.5, 2.0\}\) to find the network with the optimal variance, which was \(\sigma =
1.0\). It is possible that only adding independent Gaussian noise to one layer
is not enough to regularize the network. This motivates the need to add it to
multiple layers in order to understand its true effect.

Correlated Gaussian noise achieves the best results at classification of the
clean test set, and on five out of the six occlusion classes. The improvement
from the baseline model is \(\sim 5\%\) after being added only in the first
layer, which is encouraging. To ensure that the classification improvement was
consistent across all classes and not only for some outliers, we ran the
breakdown of the class predictions for each set shown in Table~\ref{table:classPredict1}. It can be seen that the improvements are consistent
for \(\sim 70\%\) of different output classes.

Independent Poisson noise has worse performance than the baseline model across
all sets. To investigate why this occurred, we examined the distribution of the
activations of layer 1 (samples from the Poisson distribution). The values were
highly variable in the range \([3,35]\). This may stem from the fact
that we back-propagate the mean rate of the Poisson instead of the actual
sample value, which could affect learning. We hypothesize that a penalty on the
activations, combined with the independent Poisson noise could achieve
acceptable classification results.

The correlated Poisson model also achieves better results than the baseline
model across the test set and all occlusion classes and has the best result on
the vertical half occlusion set. It also performs better
compared to the independent Poisson model across all occlusion sets showing
that correlations have an appreciable effect on the classification performance.

\subsection{Presence of Dropout} \label{yesDrop}

Dropout was applied to the input image, as well as after each of the layers that
has a stride
of 2 (simply a convolutional layer that replaces pooling
\cite{Springenberg2014}, specifically layer 3 and 6). The
dropout probability at the input was 20\% and was 50\% otherwise.

The classification performance of the models that incorporate dropout are
summarized in Table~\ref{table:recog2}, with the model abbreviations shown in
Table~\ref{table:stochasticModels}.

We observe that, when compared to the model without dropout, the baseline model
with dropout performed better across all classes with a mean
performance increase on the test set of \(\sim 10\%\).

Independent Gaussian models improve performance over the baseline model in all
sets of images. This is expected, as independent noise is
similar to dropout. Hence, it can be interpreted as adding dropout to the first
layer of the network, which can explain the performance benefit.

The correlated Gaussian models improved the performance of the network for five
out of the seven categories against the baseline model with dropout. This is
promising because the addition of dropout into the model with correlated noise
increases performance on average.

The independent Poisson model with dropout improved performance against the
baseline model with dropout on the test set and six out of the seven occlusion
sets. We hypothesized earlier that an activity penalty with an independent
Poisson model can lead to strong classification performance, but it seems
that dropout along with Poisson noise seems to be a good regularizer for the
network.

The correlated Poisson models perform better than the baseline model on all the
occluded image sets. The improvements with the correlated model are consistent
across different image classes as shown in Table \ref{table:classPredict2}.

In general, the addition of correlated noise into the baseline model with and
without dropout increases recognition performance across the occlusion sets.

\section{Conclusion}

In this work, we proposed a model of a stochastic neural network whose activity
values are correlated based on the spatial spread and selectivity of the
neurons. We tested different noise models, as described in Section~\ref{noiseFramework}, on the classification task of occluded and non-occluded
images.

A simple model of correlated variability, inspired by the visual cortex, is
modeled using the spatial distance between units in an output feature map and
the kernel weights that produced them. The variability can be drawn from
different distributions of specified correlation; for example, Gaussian and Poisson distributions.
Depending on whether the sampling function is differentiable, back-propagation
can continue as normal, or it will rely on an estimate of the gradient. There are
different methods to estimate the gradient, including re-parameterization and
straight-through estimation.

Preliminary results show that correlated variability added to a single layer
can perform better than other noise models. In fact, in ten of the twelve
occlusion
cases we tested, both with and without additional regularization, our best
performing models had some form of correlated noise. It remains to be
investigated whether this trend is robust across other network
architectures, types of occlusion, etc. It also remains to be investigated
how adding correlated
noise to multiple layers (as opposed to a single layer) affects the
classification performance of the network. However, sampling from high-dimensional correlated distributions can be computationally expensive. One of
the main areas of focus in the future will be how to balance computational
tractability with the performance benefits of incorporating correlation.

We considered Poisson distributions, as cortical neurons display Poisson-like
statistics in their spike arrival times. Samples from a Poisson distribution
can also approximate dropout in the case when the rates are low. However, we
observed that the activations in the Poisson models are highly variable.
The standard deviation of Poisson noise scales with the square root of the
activation, so we expect that Poisson noise is a weaker regularizer when the
activations are higher. This motivates the need for an activity penalty, in
addition to the Poisson noise. Our experiments show that dropout paired with
Poisson noise is also a strong regularizer, improving recognition performance
on the occluded image sets.

While we used a similar version of the straight-through estimator for the
Poisson models, there is a way to form a continuous relaxation of a Poisson
distribution using the work in \cite{Maddison2016,Jang2016}. It is also
possible to use the idea of deconvolutions/fractional strided convolutions
\cite{Zeiler2013} to understand
the effect of correlated noise in input space. We intend to pursue this as part
of our future work.

\section*{Acknowledgements}

We would like to thank Victor Reyes Osorio and Brittany Reiche for their feedback on this work.

\bibliographystyle{IEEEtran}
\bibliography{IEEEabrv,paper}

\begin{thebibliography}{10}
\providecommand{\url}[1]{#1}
\csname url@samestyle\endcsname
\providecommand{\newblock}{\relax}
\providecommand{\bibinfo}[2]{#2}
\providecommand{\BIBentrySTDinterwordspacing}{\spaceskip=0pt\relax}
\providecommand{\BIBentryALTinterwordstretchfactor}{4}
\providecommand{\BIBentryALTinterwordspacing}{\spaceskip=\fontdimen2\font plus
\BIBentryALTinterwordstretchfactor\fontdimen3\font minus
  \fontdimen4\font\relax}
\providecommand{\BIBforeignlanguage}[2]{{%
\expandafter\ifx\csname l@#1\endcsname\relax
\typeout{** WARNING: IEEEtran.bst: No hyphenation pattern has been}%
\typeout{** loaded for the language `#1'. Using the pattern for}%
\typeout{** the default language instead.}%
\else
\language=\csname l@#1\endcsname
\fi
#2}}
\providecommand{\BIBdecl}{\relax}
\BIBdecl

\bibitem{Zeiler2013}
M.~D. {Zeiler} and R.~{Fergus}, ``{Visualizing and Understanding Convolutional
  Networks},'' \emph{ArXiv e-prints}, Nov. 2013.

\bibitem{Yamins2014}
D.~L.~K. Yamins, H.~Hong, C.~F. Cadieu, E.~A. Solomon, D.~Seibert, and J.~J.
  DiCarlo, ``Performance-optimized hierarchical models predict neural responses
  in higher visual cortex,'' \emph{Proceedings of the National Academy of
  Sciences}, vol. 111, no.~23, pp. 8619--8624, 2014.

\bibitem{Zohary1994}
E.~Zohary, M.~N. Shadlen, and W.~T. Newsome, ``Correlated neuronal discharge
  rate and its implications for psychophysical performance,'' \emph{Nature},
  vol. 370, pp. 140--143, Jul 1994.

\bibitem{Wang2017}
X.~Wang, A.~Shrivastava, and A.~Gupta, ``A-fast-rcnn: Hard positive generation
  via adversary for object detection,'' \emph{CoRR}, vol. abs/1704.03414, 2017.

\bibitem{Srivastava2014}
N.~Srivastava, G.~Hinton, A.~Krizhevsky, I.~Sutskever, and R.~Salakhutdinov,
  ``Dropout: A simple way to prevent neural networks from overfitting,''
  \emph{J. Mach. Learn. Res.}, vol.~15, no.~1, pp. 1929--1958, Jan. 2014.

\bibitem{Bengio2013}
Y.~Bengio, N.~L{\'{e}}onard, and A.~C. Courville, ``Estimating or propagating
  gradients through stochastic neurons for conditional computation,''
  \emph{CoRR}, vol. abs/1308.3432, 2013.

\bibitem{Kingma2013}
D.~P. {Kingma} and M.~{Welling}, ``{Auto-Encoding Variational Bayes},''
  \emph{ArXiv e-prints}, Dec. 2013.

\bibitem{Maddison2016}
C.~J. {Maddison}, A.~{Mnih}, and Y.~{Whye Teh}, ``{The Concrete Distribution: A
  Continuous Relaxation of Discrete Random Variables},'' \emph{ArXiv e-prints},
  Nov. 2016.

\bibitem{Jang2016}
E.~{Jang}, S.~{Gu}, and B.~{Poole}, ``{Categorical Reparameterization with
  Gumbel-Softmax},'' \emph{ArXiv e-prints}, Nov. 2016.

\bibitem{Poole2014}
B.~{Poole}, J.~{Sohl-Dickstein}, and S.~{Ganguli}, ``{Analyzing noise in
  autoencoders and deep networks},'' \emph{ArXiv e-prints}, Jun. 2014.

\bibitem{DeVries2017}
T.~{DeVries} and G.~W. {Taylor}, ``{Dataset Augmentation in Feature Space},''
  \emph{ArXiv e-prints}, Feb. 2017.

\bibitem{Zhang2017-yx}
H.~{Zhang}, M.~{Cisse}, Y.~N. {Dauphin}, and D.~{Lopez-Paz}, ``{mixup: Beyond
  Empirical Risk Minimization},'' \emph{ArXiv e-prints}, Oct. 2017.

\bibitem{Smith2008}
M.~A. Smith and A.~Kohn, ``Spatial and temporal scales of neuronal correlation
  in primary visual cortex,'' \emph{Journal of Neuroscience}, vol.~28, no.~48,
  pp. 12\,591--12\,603, 2008.

\bibitem{Springenberg2014}
J.~T. Springenberg, A.~Dosovitskiy, T.~Brox, and M.~A. Riedmiller, ``Striving
  for simplicity: The all convolutional net,'' \emph{CoRR}, vol. abs/1412.6806,
  2014.

\bibitem{tensorflow2015}
M.~Abadi, A.~Agarwal, P.~Barham, E.~Brevdo, Z.~Chen, C.~Citro, G.~S. Corrado,
  A.~Davis, J.~Dean, M.~Devin, S.~Ghemawat, I.~J. Goodfellow, A.~Harp,
  G.~Irving, M.~Isard, Y.~Jia, R.~J{\'{o}}zefowicz, L.~Kaiser, M.~Kudlur,
  J.~Levenberg, D.~Man{\'{e}}, R.~Monga, S.~Moore, D.~G. Murray, C.~Olah,
  M.~Schuster, J.~Shlens, B.~Steiner, I.~Sutskever, K.~Talwar, P.~A. Tucker,
  V.~Vanhoucke, V.~Vasudevan, F.~B. Vi{\'{e}}gas, O.~Vinyals, P.~Warden,
  M.~Wattenberg, M.~Wicke, Y.~Yu, and X.~Zheng, ``Tensorflow: Large-scale
  machine learning on heterogeneous distributed systems,'' \emph{CoRR}, vol.
  abs/1603.04467, 2016.

\end{thebibliography}

\end{document}